# Analyse Comparative des Manipulateurs 3R à Axes Orthogonaux

**Maher Baili — Damien Chablat — Philippe Wenger**

*Institut de Recherche en Communications et Cybernétique de Nantes*
*UMR CNRS 6597*
*1 rue de la Noë, BP 92101, 44321 Nantes Cedex 03 France*
 *Damien.chablat@irccyn.ec-nantes.fr*

*RÉSUMÉ. Une famille de manipulateurs à 3 articulations rotoïdes et à axes orthogonaux sans décalage sur le troisième segment se divise en neuf topologies d'espace de travail différentes. Une topologie est définie par le nombre de points cusps et de nœuds qui apparaissent sur les surfaces de singularité. En se basant sur cette classification, on évalue les manipulateurs étudiés selon deux indices de performances dont le premier est le conditionnement qui est relatif à l'espace articulaire tandis que le second est la proportion de la région à 4 solutions par rapport à une sphère englobant l'espace de travail à l'espace de travail. On détermine la topologie d'espace de travail dans à laquelle se trouve appartiennent les manipulateurs ayant les meilleurs indices de performances.*

*ABSTRACT. A family of 3R orthogonal manipulators without offset on the third body can be divided into exactly nine workspace topologies. The workspace is characterized in a half-cross section by the singular curves. The workspace topology is defined by the number of cusps and nodes that appear on these singular curves. Based on this classification, we evaluate theses manipulators by the condition number related to the joint space and the proportion of the region with four inverse kinematic solutions compared to a sphere containing all the workspace. This second performance number is in relation with the workspace. We determine finally le topology of workspace to which belong manipulators having the best performance number values.*

*MOTS-CLÉS : manipulateur orthogonal, classification, singularité, cusp, nœud, topologie d'espace de travail, conditionnement.*

**1. Introduction**

Malgré les progrès technologiques, les industriels n'utilisent que des manipulateurs « standards » pour lesquels les axes des articulations successives peuvent être parallèles. Pour répondre à des besoins d'implantation particuliers, les concepteurs de sites robotisés ont été demandeurs de solutions innovantes. Par conséquent, des fabricants de robots ont été amenés à employer des morphologies originales avec porteur à axes orthogonaux. Leur mise en œuvre a fait apparaître des comportements inattendus (Hemmingson *et al.*, 1996) : L'utilisation des lois de commande usuelles conduisait à des réponses erratiques. Les roboticiens ont longtemps pris pour acquis que, pour changer de posture, il fallait franchir une singularité. Borrel pensait l'avoir démontré théoriquement en 1986 (Borrel *et al.*, 1986). Mais un contre-exemple a remis en cause cette propriété : un manipulateur 6R peut changer de posture sans passer par une singularité (Parenti, 1988). Un résultat analogue a été publié en 1991 pour des manipulateurs 3R (Burdick, 1991). De tels manipulateurs sont appelés manipulateurs cuspidaux. Une condition nécessaire et suffisante et donnée dans (El Omri, 1996), un manipulateur peut changer de posture sans franchir une singularité, si et seulement si, il existe dans son espace de travail, un point où trois solutions du modèle géométrique inverse coïncident : Un point cusp. Une condition dépendant des paramètres géométriques pour qu'un manipulateur à 3 articulations rotoïdes, à axes orthogonaux et sans décalage entre les deux derniers axes est donnée sous forme symbolique (Baili, 2004). En outre, une classifiaction selon les topologies d'espace de travail de ces manipulateurs est établie. Cette classification reste destinée à des analyses globales d'accessibilité ou de parcourabilité. Le but de ce papier est de proposer un complément de cette étude en évaluant les manipulateurs selon deux indices de performances dont l'un est relatif à l'espace articulaire et l'autre à l'espace de travail.

Ce papier est organisé comme suit. Dans la prochaine section, on présentera la famille de manipulateurs étudiée et on rappellera la classification établie dans (Baili, 2004). Dans la section 3, on analysera cette classification selon les deux indices de performances choisis. Dans la dernière section, on résumera l'apport du travail présenté dans ce papier.

**2. Préliminaires**

**2.1.** *Présentation de la famille de manipulateurs étudiée*

La figure 1 représente l'architecture cinématique des manipulateurs étudiés dans leur configuration de référence. L'effecteur représenté par le point P est repéré par ses trois coordonnées cartésiennes *x*, *y* et *z* définies dans le repère de référence (**O**,

**X**, **Y**, **Z**) attaché à la base du manipulateur. Les manipulateurs étudiés dans ce papier n'ont pas de décalage entre les deux derniers axes. Les paramètres géométriques restants à considérer sont $d_2$, $d_3$, $d_4$ et $r_2$. Les angles $\alpha_2$ et $\alpha_3$ sont égaux à -90° et 90° respectivement. On ne considèrera pas de butées articulaires, par suite, les variables articulaires $\theta_1$, $\theta_2$ et $\theta_3$ sont illimitées.

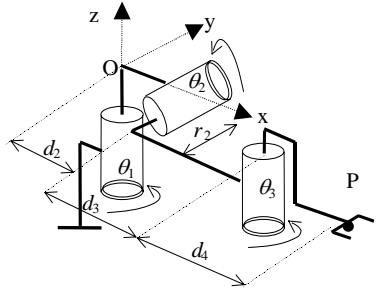

**Figure 1.** *Famille de manipulateurs étudiée*

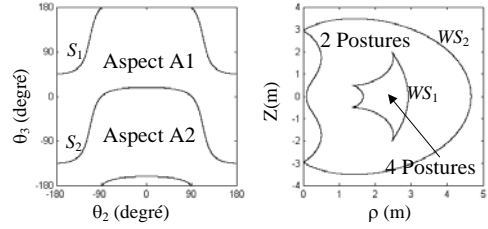

$d_2 = 1, d_3 = 2, d_4 = 1,5, r_2 = 1$

**Figure 2.** *les branches de singularités dans l'espace articulaire (à gauche) et les surfaces de singularités dans l'espace de travail (à droite)*

## 2.2. Propriétés des manipulateurs

Les manipulateurs étudiés dans cet article possèdent les propriétés suivantes :Ils peuvent être cuspidal (il peut changer de posture sans franchir une singularité) ; Il peuvent avoir ou pas une cavité toroïdale dans son espace de travail (un trou à l'intérieur de l'espace de travail) et ils peuvent être générique[1] / non générique et binaire[2] / quaternaire[3]. Ces propriétés sont directement liées à la topologie des surfaces de singularité dans l'espace de travail.

Le déterminant de la matrice Jacobienne d'un manipulateur de cette famille est donné par l'équation [1].

$$\det(\mathbf{J}) = d_4(d_3 + d_4 c_3)\left[d_2 s_3 + (d_3 s_3 - r_2 c_3)c_2\right] \quad [1]$$

où $c_i = \cos(\theta_i)$ et $s_i = \sin(\theta_i)$. Les singularités sont définies par $\det(\mathbf{J}) = 0$. Le tracé de $\det(\mathbf{J}) = 0$ forme deux ou quatre branches de singularités dans $-\pi \leq \theta_2 \leq \pi$ et $-\pi \leq \theta_3 \leq \pi$ selon que $d_3 > d_4$ ou pas.

---

[1] Un manipulateur est générique si ses branches de singularité dans l'espace articulaire ne se coupent pas
[2] Un manipulateur est binaire s'il a deux solutions au Modèle Géométrique Inverse (MGI)
[3] Un manipulateur est quaternaire s'il a quatre solutions au MGI

Lorsque $d_3 > d_4$, le premier facteur du déterminant donné par l'équation [1] ne s'annule pas, par conséquent, on obtient deux branches de singularités distinctes $S_1$ et $S_2$ dans l'espace articulaire (El Omri, 1996) qui divisent l'espace articulaire en deux domines A1 et A2 dépourvus de singularités appelés aspects (Borrel *et al.*, 1988), voir figure 2.

Lorsque $d_3 \leq d_4$, l'effecteur rencontre l'axe 2 et on a $\theta_3 = \pm \arccos(-d_3/d_4)$. Dans ces conditions, deux droites supplémentaires apparaissent dans l'espace articulaire défini tels que $-\pi \leq \theta_2 \leq \pi$ et $-\pi \leq \theta_3 \leq \pi$. Ces deux droites peuvent couper ou pas les branches de singularité $S_1$ et $S_2$ dans l'espace articulaire. Le nombre d'aspects dépend ainsi de ces intersections. Notons que dans le cas $d_3 \leq d_4$, aucune surface de singularité supplémentaire autre que la surface de singularité intérieure $WS_1$ (image de $S_1$) et la surface de singularité extérieure $WS_2$ (image de $S_2$) n'apparaissent dans l'espace de travail. Ceci s'explique par le fait que lorsque l'effecteur rencontre l'axe 2, la rotation selon $\theta_2$ n'entraîne pas un changement de la position de l'effecteur.

Le figure 2 correspond à un manipulateur cuspidal : il a 4 cusps sur la surface de singularité intérieure $WS_1$. Cette surface délimite une région dont le degré d'accessibilité est de 4 (le nombre de solutions au MGI ou le nombre de postures). Toutefois, ce manipulateur n'a par de nœuds et pas de cavité toroïdale dans son espace de travail. Sur la base de la topologie des surfaces de singularité dans l'espace de travail définie par le nombre de points singuliers particuliers qui apparaissent (les cusps et les nœuds), une classification exhaustive de l'espace des paramètres en différentes topologies d'espace de travail a été établie (Baili *et al.*, 2004). Cette classification se base sur les propriétés : générique / non générique, binaire / quaternaire, le nombre d'aspects dans l'espace articulaire, le nombre de points cusps, le nombre de nœuds et la présence ou pas de cavité toroïdale dans l'espace de travail.

### 2.3. Classification selon le nombre de cusps

L'objectif est de partitionner l'espace des paramètres en plusieurs cellules où le nombre de points cusps reste constant. Afin de diminuer le nombre d'inconnues du problème et sans perte de généralités, on normalise tous les paramètres par $d_2$. Les paramètres à considérer sont désormais $d_3$, $d_4$ et $r_2$. En utilisant des outils poussés de calcul algébrique, (Corvez *et al.*, 2002) montre que l'espace des paramètres peut se diviser en 105 cellules où le nombre de points cusps est constant. (Baili *et al.*, 2003) ont regroupé ces 105 cellules en seulement 5 domaines où les manipulateurs ne peuvent avoir que 0, 2 ou 4 cusps. Ceci montre que parmi les 5 surfaces données dans (Corvez *et al.*, 2002), une ou plusieurs ne sont pas pertinentes. Toutefois, (Baili *et al.*, 2003) ne précise pas les surfaces pertinentes parmi les 5. C'est dans (Baili *et al.*, 2004) que les équations des 4 surfaces pertinentes $C_1$, $C_2$, $C_3$ et $C_4$ ont été données sous forme symbolique dépendant des paramètres géométriques. Les équations [2], [3], [4] et [5] représentent les 4 surfaces pertinentes.

$$C_1 : d_4 = C_{1d} \text{ où } C_{1d} = \sqrt{\frac{1}{2}\left(d_3^2 + r_2^2 - \frac{(d_3^2 + r_2^2)^2 - d_3^2 + r_2^2}{AB}\right)} \quad [2]$$

$$C_2 : d_4 = C_{2d} \text{ où } C_{2d} = \frac{d_3}{1+d_3} \cdot A \quad [3]$$

$$C_3 : d_4 = C_{3d} \text{ où } C_{3d} = \frac{d_3}{d_3 - 1} \cdot B \quad \text{et} \quad d_3 > 1 \quad [4]$$

$$C_4 : d_4 = C_{4d} \text{ où } C_{4d} = \frac{d_3}{1 - d_3} \cdot B \quad \text{et} \quad d_3 < 1 \quad [5]$$

où $A = \sqrt{(d_3+1)^2 + r_2^2}$ et $B = \sqrt{(d_3-1)^2 + r_2^2}$

La figure 3 représente les 4 surfaces de séparations et les 5 domaines. Les domaines 1, 2, 3, 4 et 5 sont associés aux manipulateurs ayant 0, 4, 2, 4 et 0 cusps dans leurs espaces de travail respectivement.

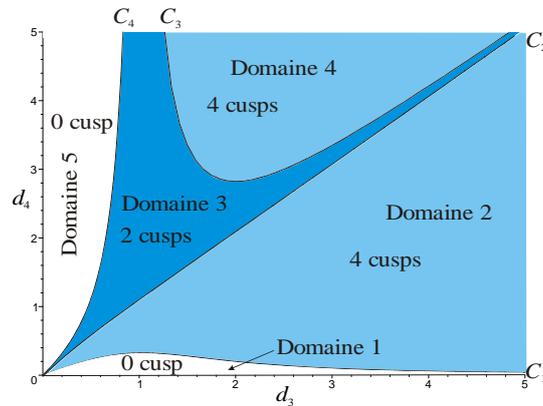

**Figure 3.** *Les* 4 *surfaces de séparations et les* 5 *domaines dans une section* ($d_3$, $d_4$) *pour* $r_2 = 1$

### 2.4. Classification selon le nombre de nœuds

On affine maintenant l'étude en déterminant toutes les topologies d'espace de travail possibles. Pour cela, on part de chaque domaine ayant un nombre de points cusps constant et on le partage en plusieurs topologies d'espace de travail. Chaque sous domaine définit une topologie d'espace de travail que l'on note $WT_i$. (Baili, 2004) montre que les manipulateurs appartenant à la famille étudiée ne peuvent avoir que 0, 1, 2, 3 ou 4 nœuds et qu'il existe exactement 9 topologies d'espace de travail différentes. Les 3 surfaces supplémentaires $E_1$, $E_2$ et $E_3$ permettant de subdiviser les 5 domaines en 9 topologies d'espace de travail correspondent aux équations [6], [7], et [8] respectivement.

$$E_1 : d_4 = E_{1d} \text{ où } E_{1d} = \frac{1}{2}(A - B) \qquad [6]$$

$$E_2 : d_4 = E_{2d} \text{ où } E_{2d} = d_3 \qquad [7]$$

$$E_3 : d_4 = E_{3d} \text{ où } E_{3d} = \frac{1}{2}(A + B) \qquad [8]$$

La figure 4 représente les 7 surfaces de séparations et les 9 topologies d'espace de travail. Dans chaque topologie d'espace de travail, on précise le nombre de cusps et le nombre de nœuds.

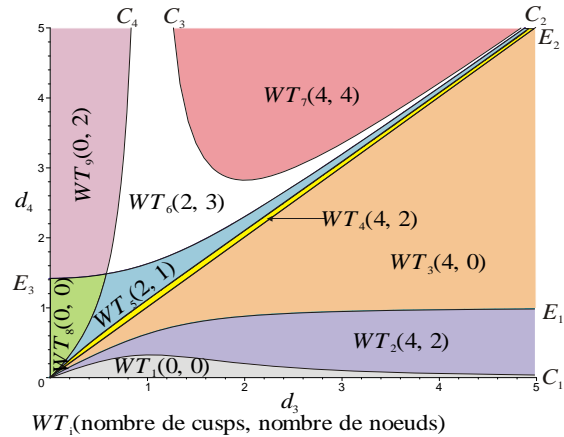

$WT_i$(nombre de cusps, nombre de noeuds)

**Figure 4.** *Les 7 surfaces de séparations et les 9 topologies d'espace de travail dans une section* $(d_3, d_4)$ *pour* $r_2 = 1$

La figure 5 représente un arbre de classification de la famille de manipulateurs étudiée en 9 topologies d'espace de travail différentes. Pour chaque topologie d'espace de travail, on indique le nombre de cusps, le nombre de nœuds, le nombre d'aspects, si le manipulateur est binaire ou quaternaire. On note que le triplet (1, 1, 0) veut dire que le manipulateur a une cavité toroïdale, une région à 2 solutions au MGI et 0 région à 4 solutions au MGI.

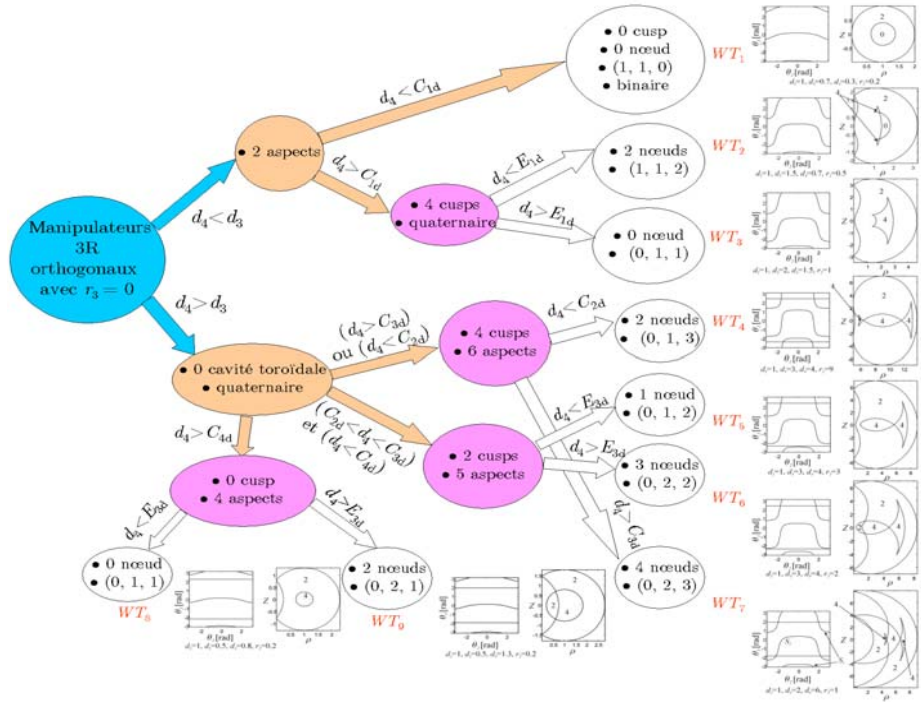

**Figure 5.** *Arbre de classification en 9 topologies d'espace de travail*

## 3. Analyse de la classification et indices de performances

La classification de la famille de manipulateurs selon leurs topologies d'espace de travail représente un outil pour l'ingénieur dans sa recherche de manipulateurs innovants. Cette classification propose des informations souvent destinées à des analyses globales d'accessibilité ou de parcourabilité. Si l'on veut compléter cette étude et aider le concepteur à affiner son choix, on se retourne vers l'évaluation des performances en un point de l'espace de travail ou en une configuration articulaire donnée. Dans cette partie, on va analyser donc la famille de manipulateurs étudiée selon deux critères de performances. Le premier critère est le conditionnement, il est relatif à l'espace articulaire. Le deuxième indice de performance est relatif à l'espace de travail, il permet de déterminer les proportions des régions à 2 et à 4 solutions au MGI et de toute la région accessible par rapport à une sphère dont le rayon est la portée du manipulateur[4].

---

[4] La porté d'un manipulateur est la distance la plus éloignée que peut atteindre son effecteur

### *3.1. Analyse selon l'espace articulaire*

Le conditionnement ou indice d'isotropie d'un manipulateur est définie comme le conditionnement de sa matrice jacobienne cinématique (Salisbury *et al.*, 1982). C'est le rapport entre la plus grande et la plus petite valeur singulière de **J** (équation [9]).

$$K = \sigma_{max} / \sigma_{min} \qquad [9]$$

Notons que cette définition est valable uniquement lorsque les éléments de la matrice jacobienne **J** ont la même dimension. Dans le cas où la tâche est définie en position et en orientation, il est possible d'homogénéiser les dimensions en divisant les lignes de **J** correspondant aux positions par « la longueur caractéristique » du manipulateur étudié (Angeles, 1997).

Le calcul du conditionnement de la matrice jacobienne repose sur la connaissance de la position articulaire ($\theta_1$, $\theta_2$, $\theta_3$) et de la position de l'effecteur (x, y, z). Pour atteindre cet objectif, deux possibilités nous sont offertes : un balayage de l'espace articulaire ou un balayage de l'espace de travail. Dans le travail présenté dans ce papier, on a opté pour un balayage dans l'espace articulaire.

Pour des raisons de présentation, on calculera l'inverse du conditionnement de la matrice jacobienne cinématique qui évolue entre 0 et 1. On le notera $K^{-1}$ pour exprimer le conditionnement inverse de la matrice jacobienne. On a choisi de calculer deux indices à partir du $K^{-1}$ : $K^{-1}$ maximum et la $K^{-1}$ moyen.

Le premier indice est un indice local qui permet de trouver un ensemble de manipulateurs possédant une configuration isotrope (Angeles, 1997) alors que le second est une mesure globale. Le travail présenté dans ce papier ne peut donner tous les résultats possibles car on est obligé de faire des coupes dans l'espace des paramètres (pour une valeur de $r_2$ donnée, par exemple). On recherchera dans chaque coupe les exemples les plus représentatifs. Ces courbes d'iso-valeurs permettent de définir l'évolution des indices vers les valeurs les plus élevées par rapport aux topologies d'espace de travail définies dans les préliminaires.

Pour calculer $K^{-1}$ maximum et $K^{-1}$ moyen, on va utiliser une technique de discrétisation de l'espace de travail. D'autres méthodes telles que l'analyse par intervalles ou Monte-Carlo auraient pu être utilisées. Notre choix a été guidé par un souci de simplicité et de rapidité de mise en œuvre.

Afin de pouvoir obtenir des résultats facilement compréhensibles, on effectue un balayage dans une section ($d_3$, $d_4$) de l'espace des paramètres pour un $r_2$ donné. Pour chaque point de la section ($d_3$, $d_4$) représentant un manipulateur, on calcule les valeurs de $K^{-1}$ moyen et maximal.

On obtient dans une section de l'espace des paramètres, des courbes d'iso-valeurs dont chacune représente les manipulateurs ayant la même valeur de $K^{-1}$ moyen ainsi que la même valeur de $K^{-1}$ maximal.

La figure 6 représente les courbes d'iso-valeurs de $K^{-1}$ moyen dans une section ($d_3$, $d_4$) de l'espace des paramètres pour $r_2 = 1$. Par ailleurs, figure 7 représente les courbes d'iso-valeurs de $K^{-1}$ maximum dans la même section ($d_3$, $d_4$).

Les flèches que l'on voit sur les sections illustrées par la figure 6 (respectivement la figure 7) représentent dans chaque topologie d'espace de travail, les évolutions vers les valeurs de $K^{-1}$ moyen (respectivement maximum) les plus élevées. Dans chaque section, la zone contenant des manipulateurs ayant une valeur de $K^{-1}$ moyen (respectivement maximum) élevée est marquée par (✱). On choisit un manipulateur dans chacune de ces zones et on calcule son $K^{-1}$ moyen et son $K^{-1}$ maximum. Les valeurs de $K^{-1}$ calculées ont une précision de $10^{-4}$. Toutefois et après arrondissement, on note deux chiffres après la virgule. Ces valeurs sont données après une vérification de leurs faibles variations par rapport à la variation du pas de calcul. En effet, si on divise le pas de calcul par $10^2$, les valeurs de $K^{-1}$ ne varient que de l'ordre de $10^{-4}$.

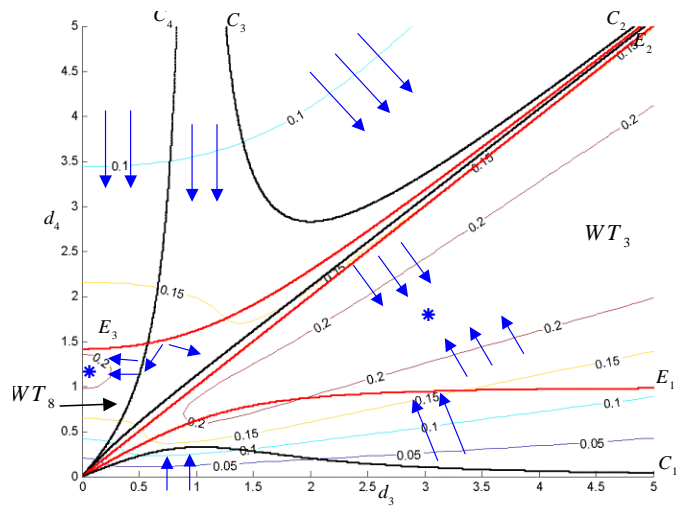

**Figure 6.** $K^{-1}$ *moyen dans une section* ($d_3$, $d_4$) *pour* $r_2 = 1$

Le manipulateur de topologie d'espace de travail $WT_3$ dont les paramètres géométriques sont : $d_2 = 1$ ; $d_3 = 3$ ; $d_4 = 1,7$ et $r_2 = 1$ a un conditionnement inverse moyen de 0,23 et un conditionnement inverse maximum de 0,81. Le deuxième manipulateur ayant une valeur de conditionnement inverse moyen élevée est de topologie d'espace de travail $WT_8$, ses paramètres géométriques sont : $d_2 = 1$ ; $d_3 = 0,1$ ; $d_4 = 1,2$ et $r_2 = 1$. son conditionnement inverse moyen est de 0,21 tandis que son conditionnement inverse maximum est de 0,83.

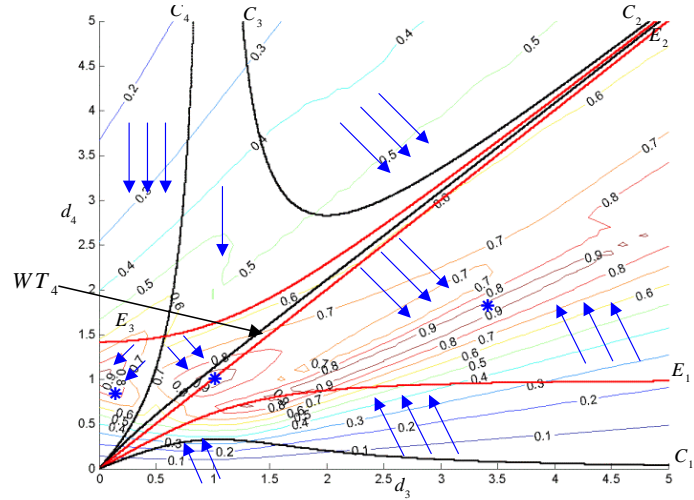

**Figure 7.** $K^{-1}$ *maximum dans une section*$(d_3, d_4)$ *pour* $r_2 = 1$

En analysant la figure 7, on remarque que les manipulateurs ayant valeur de conditionnement inverse maximum élevée appartiennent aux topologies d'espace de travail $WT_3$ et $WT_8$. On remarque aussi qu'un troisième manipulateur situé sur la surface $E_2$ a une valeur de conditionnement maximum élevée. En calculant cette valeur, on s'aperçoit qu'elle atteint la valeur maximale de 1 d'où l'isotropie de ce manipulateur. Toutefois son conditionnement inverse moyen n'est que de 0,18.

### 3.2. Analyse selon l'espace de travail

Cet indice permet, pour un manipulateur donné, de déterminer la proportion du volume de la région à 4 solutions accessible par l'effecteur par rapport au volume d'une sphère de rayon $\rho_{max}$ (représentant la portée du manipulateur) et centrée sur l'origine. Cette sphère englobe tout l'espace de travail. Dans le cas des manipulateurs étudiés dans ce papier et par une analyse géométrique simple, la portée du manipulateur est définie par l'équation [10].

$$\rho_{max} = d_4 + \sqrt{r_2^2 + (d_3+1)^2} \qquad [10]$$

Pour des raisons de symétrie de l'espace de travail, le balayage sera effectué dans la section $(\rho, z)$ et pour des valeurs de $\rho$ et $z$ positives.

On obtient une section de l'espace des paramètres représentant des courbes d'iso-valeurs de manipulateurs ayant la même proportion du volume de la région à 4 solutions par rapport à la sphère englobant l'espace de travail.

La figure 8 représente, dans une section ($d_3$, $d_4$) pour $r_2 = 1$, les proportions de la région à 4 solutions au MGI par rapport à la sphère englobant l'espace de travail.

De la même manière, on utilise des flèches pour montrer les évolutions dans chaque topologie d'espace de travail vers les manipulateurs dont la proportion de la région à 4 solutions est la plus importante.

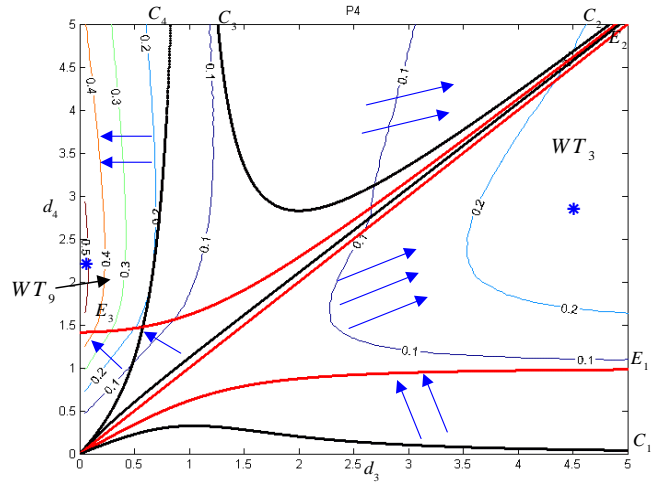

**Figure 8.** *Proportion de la région à 4 solutions dans une section($d_3$, $d_4$) pour $r_2 = 1$*

Le premier manipulateur ayant une proportion de région à 4 solutions importante appartient à la topologie d'espace de travail $WT_3$. Ses paramètres géométriques sont : $d_2 = 1$ ; $d_3 = 4,5$ ; $d_4 = 2,9$ et $r_2 = 1$. La valeur de sa proportion de la région à 4 solutions est de 0,26, celle de la région à 2 solutions est de 0,48 ce qui donne une proportion de la région accessible de 0,74 par rapport à la sphère englobant l'espace de travail.

Le second manipulateur est de topologie d'espace de travail $WT_9$. Le manipulateur dont les paramètres géométriques sont : $d_2 = 1$ ; $d_3 = 0,1$ ; $d_4 = 2,25$ et $r_2 = 1$ a une proportion de la région à 4 solutions de 0,49 et une proportion de la région à 2 solutions de 0,11. Par conséquent, la proportion de la région accessible par l'effecteur est de 0,6.

## 4. Conclusions

Les manipulateurs étudiés se classifient en 9 topologies d'espace de travail différentes. Bien que cette classification soit d'une aide précieuse pour un ingénieur recherchant des manipulateurs innovants, elle ne propose que des informations destinées à des analyses globales d'accessibilité ou de parcourabilité. Afin de la compléter, on analyse ces manipulateurs selon deux critères de performances. Le premier est relatif à l'espace articulaire et se résume à calculer les $K^{-1}$ moyen et maximum ; on a remarqué que les manipulateurs ayant les valeurs les plus élevées appartiennent aux topologies d'espace de travail $WT_3$ et $WT_8$ pour $K^{-1}$ moyen et aux topologies $WT_3$, $WT_4$ et $WT_8$ pour $K^{-1}$ maximum. Le second indice est relatif à l'espace de travail et permet de calculer la proportion de la région à 4 solutions au MGI par rapport à une sphère dont le rayon est égal à la portée du manipulateur. Pour ce critère, les topologies d'espace de travail $WT_3$ et $WT_9$ contiennent les manipulateurs ayant les meilleures proportions. En résumé, les manipulateurs ayant les meilleurs indices de performances appartiennent nécessairement à la topologie d'espace de travail $WT_3$.

## 5. Bibliographie